\documentclass{article}

\usepackage[numbers]{natbib}

\usepackage[preprint]{neurips_2024}

\usepackage[utf8]{inputenc} 
\usepackage[T1]{fontenc}    
\usepackage{hyperref}       
\usepackage{url}            
\usepackage{booktabs}       
\usepackage{amsfonts}       
\usepackage{nicefrac}       
\usepackage{microtype}      
\usepackage{xcolor}         

\usepackage{amsmath}
\usepackage{graphicx}
\usepackage{booktabs}
\usepackage{multirow}

\newcommand*{\affaddr}[1]{#1} 

\newcommand*{\email}[1]{\texttt{#1}}

\title{TART: Token-based Architecture Transformer for Neural Network Performance Prediction}

\author{%
    Yannis Y.~He
    \thanks{
        Equally advised by Graham W. Taylor, affiliated with Vector Institute, University of Guelph (\href{mailto:graham.taylor@vectorinstitute.ai}{graham.taylor@vectorinstitute.ai}) and Boris Knyazev, affiliated with Samsung - SAIT AI Lab (\href{mailto:borknyaz@gmail.com}{borknyaz@gmail.com})
    } \\ \\
  \affaddr{Vector Institute} \hspace*{0.02 \linewidth}
  \affaddr{University of Toronto} \\ \\
  \email{
      \href{mailto:yannis.he@mail.utoronto.ca}{yannis.he@mail.utoronto.ca}
      }\\ \\
  \color{violet}\url{ https://github.com/yAya-yns/TART}
}

\begin{document}

\maketitle

\begin{abstract}
    In the realm of neural architecture design, achieving high performance is largely reliant on the manual expertise of researchers. Despite the emergence of Neural Architecture Search (NAS) as a promising technique for automating this process, current NAS methods still require human input to expand the search space and cannot generate new architectures. This paper explores the potential of Transformers in comprehending neural architectures and their performance, with the objective of establishing the foundation for utilizing Transformers to generate novel networks. We propose the Token-based Architecture Transformer (TART), which predicts neural network performance without the need to train candidate networks. TART attains state-of-the-art performance on the DeepNets-1M dataset for performance prediction tasks without edge information, indicating the potential of Transformers to aid in discovering novel and high-performing neural architectures.
\end{abstract}

\section{Introduction}
Manual expertise of researchers is often the key to achieving high performance in neural architecture design, encompassing metrics such as accuracy, fairness, robustness, calibration, interpretability, latency, and memory~\cite{hutter}. However, this approach is limited by the researcher's prior experience and involves manual trial-and-error.

Neural Architecture Search (NAS) aims to automate the discovery of high-performing neural architectures with minimal human intervention~\cite{ren_xiao_chang_huang_li_chen_wang_2021}. Conventional NAS methods often use directed acyclic graphs (DAG)~\cite{liu_simonyan_yang_2019} or a sequence of operators~\cite{luo_tian_qin_chen_liu_2019} to represent a neural architecture. However, the strong token-processing capability of Transformers~\cite{attentionIsAllYouNeed} provides an incentive for representing neural architectures as tokens. Moreover, Transformers have illustrated promising results in graph learning~\cite{kim_nguyen_min_cho_lee_lee_hong_2022} as well as understanding token-based content, such as texts~\cite{zhang_song_li_zhou_song_2022} and music notes~\cite{huang_vaswani_uszkoreit_shazeer_simon_hawthorne_dai_hoffman_dinculescu_eck_et_al._2018}. Although prior research discovered that the token-based graph representation could achieve state-of-the-art performance in chemical compound structure generation~\cite{krenn_häse_nigam_friederich_aspuru-guzik_2020}, as far as our knowledge extends, no comprehensive research has been conducted on Transformers for understanding token-based neural architecture representation.  

\begin{figure}
  \centering
  \includegraphics[width=0.8\textwidth]{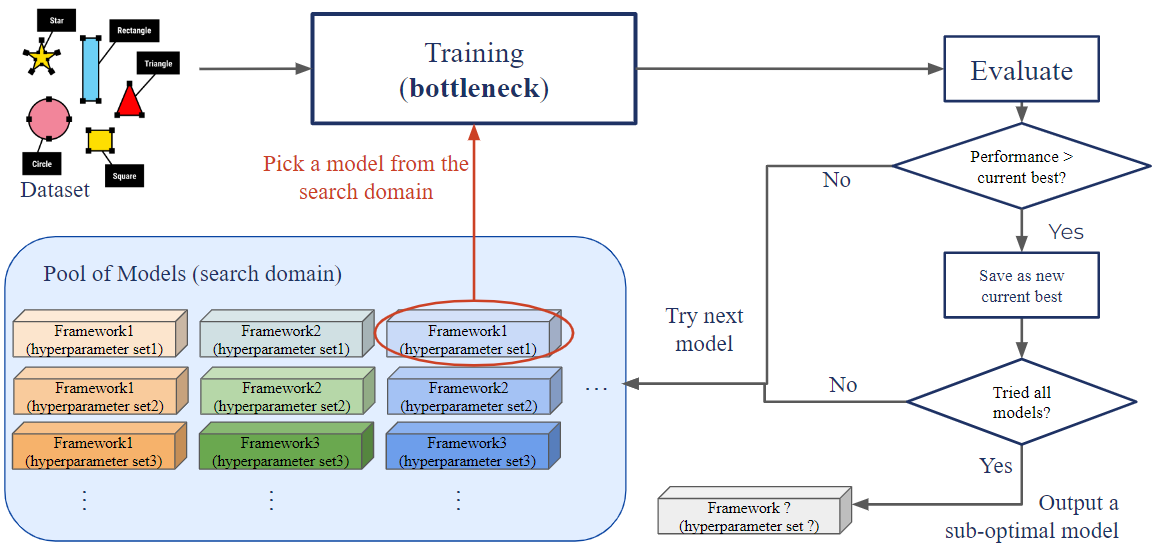}
  \caption{Current Neural Architecture Search process.}
  \label{current_nas}
\end{figure}

Most of the NAS algorithms follow a typical search process (Fig.~\ref{current_nas}). \textit{Step 1) Define the search space}: Given a task, researchers need to manually select candidate networks from all existing architectures to form a pool of models, also known as the search domain. This can include the type and number of layers, the type of activation functions, the number of neurons in each layer, and other architectural hyperparameters. \textit{Step 2) Select a candidate network}: There are various strategies for exploring the search domain, such as random search, evolutionary algorithms, reinforcement learning, and gradient-based optimization methods. \textit{Step 3) Train and evaluate candidate architectures}: For each architecture, it is trained on a subset of the training data and evaluated on a validation set. The evaluation metric can be accuracy, loss, F1-score, or any other metric that measures the performance of the model on the task at hand. \textit{Step 4) Update search strategy}: Based on the performance of the candidate architectures, the search strategy is updated to explore more promising regions of the search space. \textit{Step 5) Repeating steps 2-4}: The process of searching, training, evaluating, and updating the search strategy is repeated until the best architecture is found, or a stopping criterion is reached. 

\paragraph{Motivations} There are two major disadvantages to the current NAS process: 1) high search costs and 2) sub-optimal search outcomes. Despite improvement compared with the traditional trial-and-error approach, training the huge number of possible architectures in the search domain prior to performance evaluation is time and computational resources-consuming. Moreover, matching an existing model that has been designed for a different task or dataset may not be suitable for a new task. Current NAS approaches still rely on manual work to expand the search domain and lack the capability to generate brand-new architectures. 

This research aims to reduce the computational cost of neural architecture search by using a predictive model to estimate the performance of new, unseen candidate architectures without actually training the models; thus eliminating the time-consuming training and evaluation process that is typically the bottleneck in the search process.
We additionally investigate the potential of Transformers~\cite{attentionIsAllYouNeed} in learning a model's performance, thus paving the way for utilizing Transformers to create novel networks.

\paragraph{Contributions} Our contributions can be summarized as:
\begin{itemize}
    \item We propose a novel approach to neural network performance prediction, referred to as TART (Token-based Architecture Transformer). To the best of our knowledge, our research group is the first to investigate the potential of Transformers in learning architecture performance by converting neural networks from computational graphs into tokens.
    
    \item We achieve state-of-the-art performance on the DeepNets-1M dataset~\cite{ppuda} for performance prediction tasks without using the edge information

    \item Our results demonstrate that the tokenization of neural architectures enhances the ability of Transformers in gaining a deeper understanding of a model's performance, which lays the groundwork for future research that utilizes Transformers to generate new neural architectures for tasks that are yet to be encountered.
    
\end{itemize}

\section{Background}

In this section, we briefly review research in NAS and parameter prediction. We also discuss representing neural architectures as graphs and representing graphs as tokens, which are critical intermediate steps to our work. Although the focus of the present study does not encompass the phase of architecture generation, a brief overview of Transformer-based generative models is also included as a means of contextualizing the incentives behind our exploration of Transformers.

\subsection{Neural Architecture Search}
NAS aims to create a neural architecture that maximizes performance with minimal human involvement in an automated way. Many architecture search algorithms are computationally demanding due to a large number of architecture evaluations. An inherent cause of inefficiency for those approaches is that architecture search is treated as a black-box optimization problem over a discrete domain. DARTS~\cite{DARTS} was created to relax the search space to be continuous so that the architecture can be optimized with respect to its validation set performance by gradient descent. The data efficiency of gradient-based optimization, as opposed to inefficient black-box search, allows DARTS to achieve competitive performance using orders of magnitude fewer computation resources. NAO~\cite{NAO} uses a similar approach by converting neural networks into continuous embedding. Once the encoder maps the discrete neural architecture into the continuous representation, the performance predictor is taken as the optimization goal of the gradient ascent. The continuous representation of the best neural architecture can be obtained by maximizing the output of the performance predictor. Lastly, a decoder is applied to convert the continuous representation back to the final discrete architecture. While various research has been conducted to improve the architecture searching process, the computational bottleneck of NAS remains as training each candidate architecture to converge. Rather than discarding the weight of a child model after each performance measurement, ENAS~\cite{ENAS} forces all child models to share weights to avoid training each child model from scratch. Although the weight sharing strategy was quickly recognized by many researchers~\cite{SMASH, AutoGAN, ContinualAndMultiTask} due to its efficiency, according to recent studies~\cite{evaluate_NAS}, this strategy may result in an incorrect evaluation of candidate architectures and difficulty in optimization. 

\subsection{Parameter and Performance Prediction}
An alternative approach to avoid training every candidate architecture is utilizing parameter prediction. Knyazev et al.~\cite{ppuda} proposed a framework that can predict parameters for diverse and large-scale architectures in a single forward pass in a fraction of a second via graph hyper network. Alongside parameter prediction, another approach to pass around the training bottleneck is utilizing performance prediction. White et al.~\cite{performance_predictor} analyzed 31 performance predictors, ranging from learning curve extrapolation to weight-sharing, supervised learning, and zero-cost proxies on four different search spaces. While most predictors are domain-specific, the study provides recommendations for performance predictors under different run-time constraints. Wei et al.~\cite{NeurPred} proposed a Neural Predictor framework, which is capable of predicting the properties, such as accuracy, inference speed, and convergence speed, given an unseen architecture and dataset. These approaches have the potential to significantly decrease the computational cost of NAS and enable more creative search methods.

\subsection{Graph Representation of Neural Architectures}
A graph is a commonly used data structure across various fields. As deep learning continues to advance, a number of research has been devoted to representing neural networks using graph structures. Thost et al.~\cite{DAG_GNN} verified the efficiency of representing neural networks as directed acyclic graphs, which were widely used in many NAS frameworks~\cite{ppuda, DARTS, GHN}. There are two types of graph representation in NAS: node-based representation~\cite{ppuda, ENAS, GHN} and edge-based representation~\cite{NAS-Bench-Graph, DARTS}. The node-based representation is where nodes represent operations, such as convolution, pooling, and edges represent connectivity between operations, i.e. forward pass flow; whereas the edge-based representation is that each node is a latent representation (e.g. a feature map in convolutional networks) and each directed edge is associated with some operations. 

We follow the node-based representation from~\cite{GHN} and define a directed acyclic computational graph as $\mathcal{A} = (\mathcal{V, E})$, where each node $v \in \mathcal{V}$ has an associated computational operator $f_v$ parameterized by $w_v$, which produces an output activation tensor $x_v$. Edges $e_{u \rightarrow v} = (u, v) \in \mathcal{E}$ represent the flow of activation tensors from node $u$ to node $v$. $x_v$ is computed by applying its associated computational operator on each of its inputs and taking summation as follows: 
\begin{equation}
    x_v = \sum_{e_{u \rightarrow v} \in \mathcal{E}}^{} f_v(x_u; w_v), \forall v \in \mathcal{V}
\end{equation}

Graph representation allows models, such as GNNs~\cite{GNN}, to learn graph structures for different datasets. In addition, the graph representation of neural networks provides opportunities to utilize graph tokenizers in order to represent neural architecture as tokens.

\subsection{Token Representation of Graphs} \label{token-rep-subsection}
In order to leverage the benefits of Transformers, which include their predictive and generative capabilities, as well as their potential to comprehend the connections between components, we studied the concept of representing a graph as a sequence of tokens. Krenn et al.~\cite{SELFIES} proposed a SELF-referencing embedded strings (SELFIES) representation of chemical molecular structures using deep learning, such as VAE~\cite{VAE} and GAN~\cite{GAN}, and demonstrated the robustness of the token-based representation. In addition, Kim et al.~\cite{pure} proposed a tokenized graph Transformer (TokenGT) that learns chemical structures as graphs in token representation via standard Transformers without graph-specific modification. In their studies, a graph's nodes and edges are treated as independent tokens. To avoid graph connectivity being discarded, they augment tokens with orthonormal node identifiers and trainable type identifiers and prove their approach is at least as expressive as an invariant graph network~\cite{2-IGN}, which is better than Graph Neural Network~\cite{GNN}. In TokenGT, two specific tokenization methods are discussed: Orthogonal Random Features (ORF) and Laplacian Eigenvector (LAP). ORF initializes a random orthogonal matrix via QR decomposition of a random Gaussian matrix. This approach requires the Transformer to recognize graph structure from the incidence information provided by node identifiers, which was proven in TokenGT to be achievable. LAP uses the Laplacian eigenvectors by performing eigendecomposition on the adjacency matrix.

Our study employs the LAP tokenization presented by TokenGT. Our decision is based on two primary considerations. Firstly, LAP demonstrated superior performance compared to the ORF tokenization in TokenGT's experimental analysis. Secondly, ORF's initialization process, which follows a normal distribution, encounters issues with sparse matrices as numerous features that could potentially be initialized as zeros. Consequently, convergence is more challenging to achieve when utilizing ORF as opposed to LAP.

\subsection{Transformer-based Generative Model}
Transformer~\cite{attentionIsAllYouNeed} was known for its generative ability~\cite{GANformer} in various domains~\cite{TransGAN, STransGAN, StyleSwin}. In the graph generation domain, besides the studies mentioned in Section \ref{token-rep-subsection}, Khajenezhad et al. developed Gransformer~\cite{Gransformer} that extends the basic autoregressive Transformer encoder to utilize the structural information of the given graph. The attention mechanism is adjusted to take into account the existence or absence of connections between each pair of nodes. In the token generation domain, Zeng et al.~\cite{TokenGAN} offered a viewpoint for obtaining image synthesis by treating it as a visual token generation problem and proposed TokenGAN, which has the ability to control image synthesis by assigning styles to content tokens through attention mechanism via Transformer.

\section{Method}
\begin{figure}
  \centering
  \includegraphics[width=\textwidth]{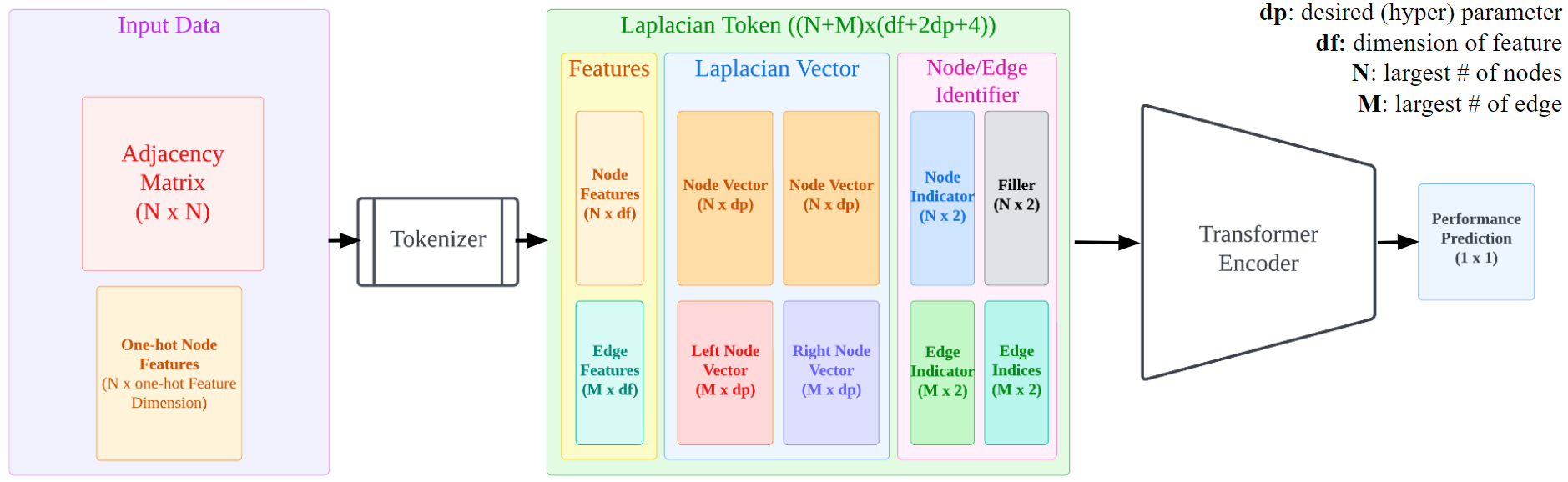}
  \caption{TART is an end-to-end neural predictor, which has three basic stages: 1) tokenization stage, 2) transformer learning stage, and 3) prediction stage.}
  \label{tart_pipeline}
\end{figure}

Our proposed TART architecture (Fig.~\ref{tart_pipeline}) is comprised of three fundamental stages: 1) tokenization, 2) transformer learning, and 3) prediction. Our primary aim in formulating this structure is not to optimize predictive capacity, but rather to demonstrate two principal objectives: 1) that Transformers can effectively learn a model's performance and 2) that tokenizers can enhance the learning capabilities of Transformers. As such, we have intentionally retained the design of each module consistent with that of related works, without any modification for the prediction task, in order to facilitate an equitable comparison.

As we discussed in the previous section, we construct our tokenizer using LAP method from TokenGT~\cite{pure} to tokenize a neural architecture from a graph representation into a Laplacian token. The token is a $(N+M) \times (d_f + 2d_p + 4)$ matrix, where $N$ is the largest number of nodes, $M$ is the largest number of edges, $d_f$ is the dimension of features, and $d_p$ is the dimension of the Laplacian Eigenvector of our choice. Instead of representing the node feature using one-hot encoding, like DeepNets-1M dataset, which we will introduce in Section~\ref{DeepNets-1M_dataset}, we choose $d_f=1$ and represent different features using a single number to reduce the size of the token. We also choose $d_p=3$ to be consistent with TokenGT.
To construct the token, we need information about the node and edges. An adjacency matrix is provided for determining connections, and a one-hot encoded node feature matrix of $N \times 15$ is also provided, where $15$ is the size of each one-hot node feature representation in the DeepNets-1M dataset. Specifically, in DeepNets-1M dataset, our tokenizer efficiently reduces the input size from about 370k parameters to around 39k.

In the Laplacian token, each row can be described in three parts: the first $d_f$ element(s) is the node features or edge features. In our case, both nodes and edges have $d_f=1$. Following the features, the second part of the token is two rows from $\mathbf{P}$ concatenated horizontally, where $\mathbf{P} \in \mathbb{R}^{N \times d_p}$ comes from computing the Laplacian Eigenvectors from the adjacency matrix of the graph. $d_p$ is the dimension of eigenvalues of our choice, which we design $d_p = 3$. For each edge $(u, v)$, $\mathbf{P}[u]$ is concatenated with $\mathbf{P}[v]$; for each node $n$, $\mathbf{P}[n]$ is concatenated with itself to match the size of $\mathbf{P}[v]$. Finally, the last $4$ elements are the node/edge identifier. The former $2$ elements out of the $4$ are one-hot encoding of whether the row represents an edge or a node, where $[0, 1]$ represents a node and $[1, 0]$ represents an edge; the latter $2$ elements are mainly used to identify an edge, in which $[-1. -1]$ is used for all nodes as fillers and $[u, v]$ is used for each edge $(u, v)$. 

We choose the design of our Transformer module to be the same as the Transformer encoder from the work proposed by Kim et al.~\cite{pure}Specifically, we have exclusively employed the encoder portion of the Transformer, given that our investigation is not directed towards network generation, rather than the classic encoder-decoder structure proposed by Vaswani et al.~\cite{attentionIsAllYouNeed}. To demonstrate the Transformer's potential for learning neural architecture, we have simply added a Fully Connected Layer after the encoder, which maps the dimensions of the Transformer output to a singular numerical value representing performance prediction.

\section{Experiments and Results}

\subsection{Datasets: DeepNets-1M}
\label{DeepNets-1M_dataset}
\begin{figure}
    \centering
    \includegraphics[width=\textwidth]{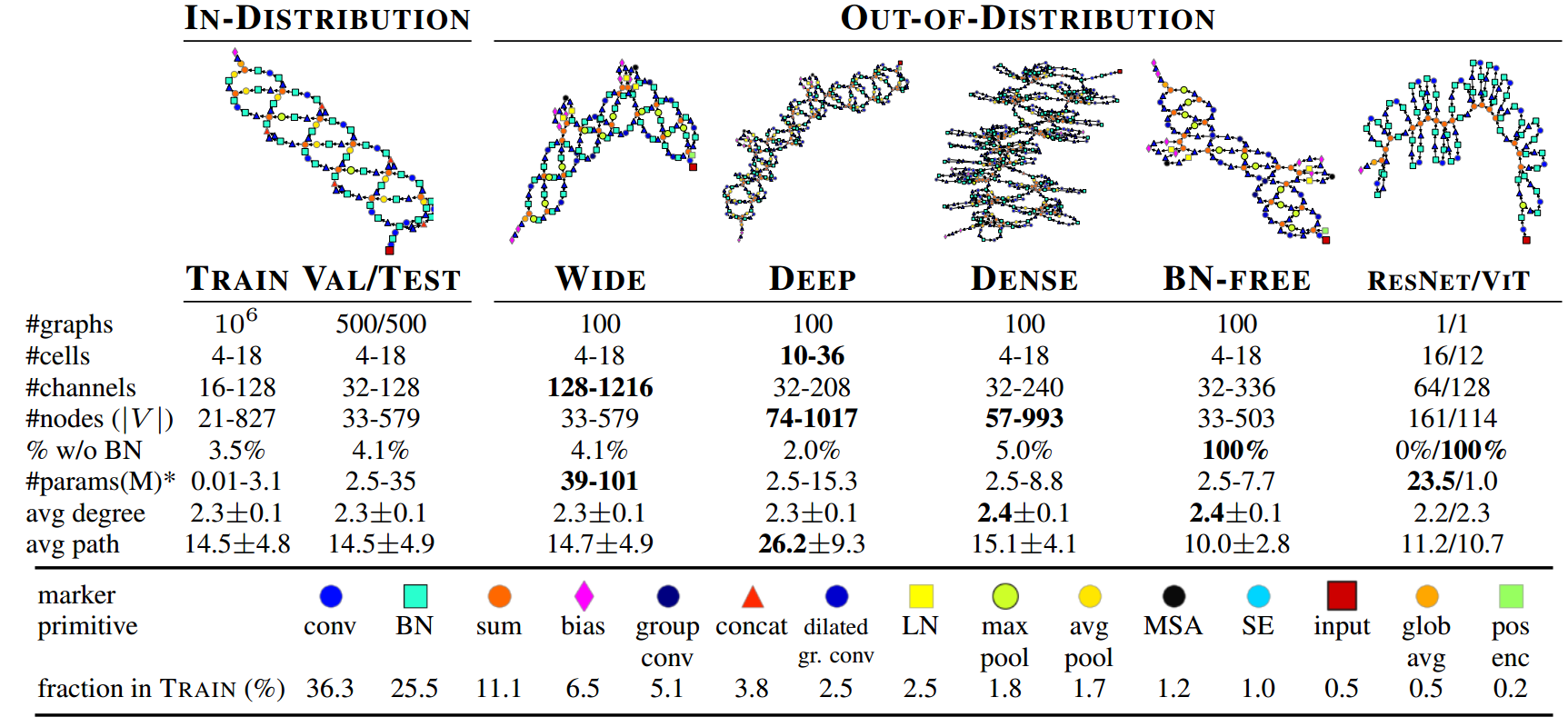}
    \caption{Examples of computational graphs (visualized using NetworkX~\cite{NetworkX}). In the visualized graphs, a node is one of the 15 primitives coded with markers shown at the bottom, where they are sorted by the frequency in the training set.}
    \label{DeepNet1M}
\end{figure}

The proposed predictor is evaluated on DeepNets-1M~\cite{ppuda} in order to compare with the baseline~\cite{NeurPred}. The DeepNets-1M dataset (figure~\ref{DeepNet1M}) is composed of neural network architectures represented as graphs where nodes are operations (convolution, pooling, etc.) and edges correspond to the forward pass flow of data through the network. While the dataset comprises 1 million neural architectures, only 1,000 of them possess performance labels. Consequently, we have restricted our training and testing splits to these 1,000 architectures, with 500 used for training and the remaining 500 for testing.

The performance labels for each of the 1,000 networks were acquired by training and evaluating them on the CIFAR-10 dataset~\cite{cifar10}. Specifically, Knyazev et al.~\cite{ppuda} measured the accuracy of each network on clear and noisy images, as well as the inference time and convergence time.

\subsection{Experiment Design}
The experimental process entails the following steps: Firstly, a predictor is trained on the 500 samples of the training split from the DeepNets-1M dataset to learn the relationship between neural architectures and their performance. Secondly, the trained predictor takes in 500 unseen and untrained architectures from the test split and predicts their performance. Then, the predicted performance is compared against the ground truth performance, which is evaluated directly after training the architecture on the CIFAR-10 dataset. The Kendall-Tau correlation~\cite{kendal_tau} is utilized as the metric to gauge the effectiveness of the predictor.

With our objective in mind, we design two sets of experiments, described in Section~\ref{experiment:PureTransformer} and~\ref{experiment:Tokenizer_Transformer}  

All results can be found in table~\ref{tab_exp_data}

\begin{table}
\caption{Kendall-Tau correlation between predicted performance and Measured performance on CIFAR-10. Each measurement is calculated by averaging 5 trials with different random seeds. Due to the computational resources limit, we are not able to train more than 30 epochs for TART. Thus, we have also train a 30 epoch pure-Transformer to compare the effects of the tokenizer}
\label{tab_exp_data}

\resizebox{\textwidth}{!}{
\begin{tabular}{@{}lccccccc@{}}
\toprule
\multicolumn{1}{c}{\textbf{Methods}} &
  \textbf{n\_layer} &
  \textbf{\begin{tabular}[c]{@{}c@{}}\# of \\ Epoch\end{tabular}} &
  \textbf{\begin{tabular}[c]{@{}c@{}}Use edge\\ features\end{tabular}} &
  \multicolumn{4}{c}{\textbf{Kendall-Tau correlation}} \\
\multicolumn{1}{c}{\textbf{}} &
  \textbf{} &
  \textbf{} &
  \textbf{} &
  \begin{tabular}[c]{@{}c@{}}Clean image\\ accuracy\end{tabular} &
  \begin{tabular}[c]{@{}c@{}}Noisy image\\ accuracy\end{tabular} &
  \begin{tabular}[c]{@{}c@{}}Inference\\ speed\end{tabular} &
  \begin{tabular}[c]{@{}c@{}}Convergence\\ speed\end{tabular} \\ \midrule
Neural Predictor                                                                    & N/A & 300 & True  & 0.482 & 0.451 & 0.695 & 0.395 \\ \midrule
pure-Transformer                                                                    & 6   & 150 & False & 0.494 & 0.458 & 0.911 & 0.494 \\
pure-Transformer                                                                    & 6   & 300 & False & 0.515 & 0.471 & 0.912 & 0.515 \\
pure-Transformer                                                                    & 12  & 300 & False & 0.544 & 0.463 & 0.913 & 0.544 \\ \midrule
pure-Transformer                                                                    & 6   & 30  & False & 0.210 & 0.137 & 0.893 & 0.210 \\
\begin{tabular}[c]{@{}l@{}}Tokenized Architecture\\ Transformer (TART)\end{tabular} & 6   & 30  & True  & 0.266 & 0.307 & 0.885 & 0.266 \\ \bottomrule
\end{tabular}
}
\end{table}

\subsection{Experiment 1: Pure-Transformer Predictor}
\label{experiment:PureTransformer}
To explore whether Transformers can learn the performance of neural architectures, we conducted an experiment in which we trained a basic Transformer encoder. Since there is no obvious way to combine and encode node features and edge features without some form of tokenizers, the Transformer is designed to only take in node features and not use any information from the adjacency matrix.

\begin{figure}
    \centering
    \includegraphics[width=\textwidth]{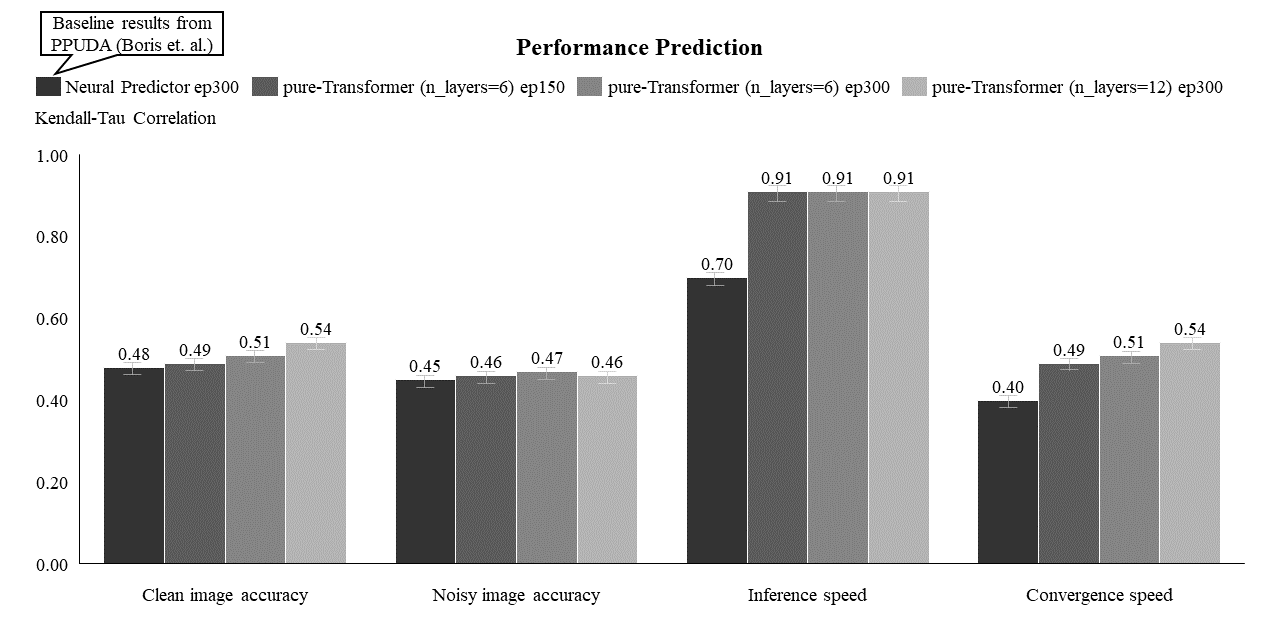}
    \caption{Measuring correlation between the predicted and ground-truth performance of models on CIFAR-10}
    \label{exp1-1_results}
\end{figure}

The results (Figure~\ref{exp1-1_results}) of the pure-Transformer predictors demonstrate the effectiveness of Transformers. Despite only utilizing node features and not using any information from the adjacency matrix, all Transformers showed higher predictive capability than the neural predictor baseline~\cite{NeurPred}, which uses both edge and node features. 

\begin{figure}
    \centering
    \includegraphics[width=0.8\textwidth]{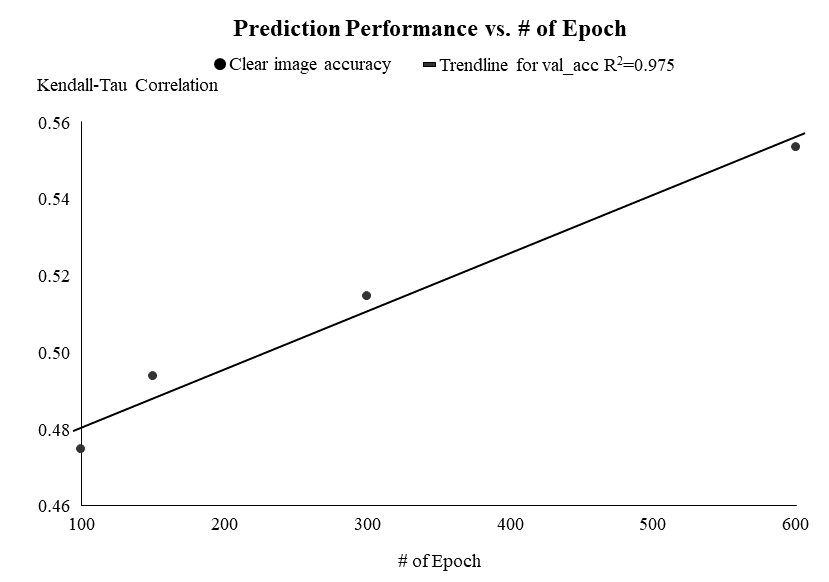}
    \caption{Although we had to halt training due to limited computational resources, our analysis of the linear regression of performance growth suggests that the predictor had not yet overfit the data.}
    \label{exp1-2_results}
\end{figure}

Although we had to halt further training of the pure-Transformer due to limited computational resources, we also observed that we had not yet overfit the pure-Transformer (Figure~\ref{exp1-2_results}). This suggests that we may be able to achieve higher performance even before making any modification to the Transformer specifically for this prediction task.

\subsection{Experiment 2: Token-based Transformer Predictor}
\label{experiment:Tokenizer_Transformer}
\begin{figure}
    \centering
    \includegraphics[width=\textwidth]{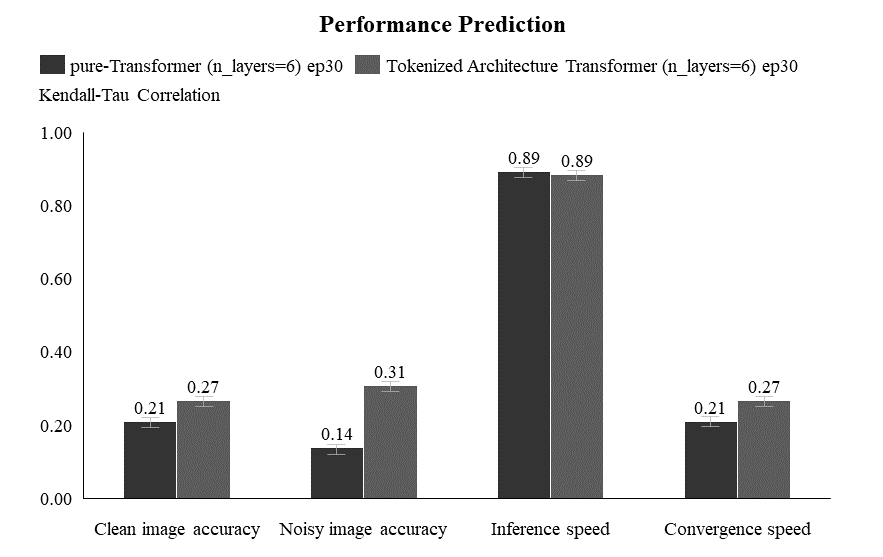}
    \caption{The performance prediction of the transformer with/without the tokenizer. The evaluation was conducted over 30 epochs.}
    \label{exp2_results}
\end{figure}

To investigate whether tokenizers can improve the Transformer's performance, we trained a complete TART architecture and compared its performance to that of a pure-Transformer. Our empirical analysis (Figure~\ref{exp2_results}) demonstrates the efficacy of incorporating a tokenizer in the training of Transformer-based predictors. Given our limited computational resources, we were only able to train the TART model for 30 epochs. To ensure a fair comparison, we evaluated its performance against that of the pure-Transformer, which was also trained for 30 epochs.

By comparing the performance of our TART architecture to that of a pure-Transformer model, we verify the positive impact of tokenization on enhancing the Transformer's ability to capture the relationship between neural architectures and their corresponding performance. This is somewhat expected because the tokenization process encodes the connections, which can be viewed as a generalization of sinusoidal positional embeddings of transformers. 

\section{Conclusions and Future Work}
In this paper, we propose TART (Token-based Architecture Transformer), a novel approach to neural network performance prediction. To the best of our knowledge, we are the first to investigate the potential of Transformers in learning architecture performance by converting neural networks into tokens. Our approach achieves state-of-the-art performance on the DeepNets-1M dataset for performance prediction tasks without using edge information.

Our results demonstrate that tokenization of neural architectures enhances the ability of Transformers to gain a deeper understanding of a model's performance. This lays the groundwork for future research that utilizes Transformers to generate new neural architectures for tasks that are yet to be encountered. Our approach offers an alternative to traditional methods of neural network performance prediction and opens up new avenues for future research.

In the future, we plan to explore several avenues to improve our TART approach. Firstly, we will investigate ways to speed up the training process of TART. The bottleneck is in the tokenization process. Converting edge features from adjacency matrix is currently done through a single-thread for-loop. One possible avenue is to preprocess the input data and vectorize the tokenization process, which is currently not. Additionally, we plan to explore the use of multi-thread tokenizers and longer training times for both the tokenizer and transformer.

Secondly, we aim to finetune the transformer and tokenizer by experimenting with different hyperparameters and architectures to further improve the performance of our predictor.

Thirdly, we will also investigate how our approach can be applied to the task of neural network generation. By training the transformer to learn the relationship between architecture and performance, we may be able to generate new neural architectures that exhibit better performance for specific tasks.

Overall, our work demonstrates the potential of using Transformers and tokenization techniques in the field of neural architecture search and performance prediction. We believe that further research in this area has the potential to lead to significant advancements in the development of more efficient and effective neural networks for various applications.

\medskip

\bibliographystyle{unsrt}
\bibliography{main}

\end{document}